\documentclass[runningheads]{llncs}
\usepackage{graphicx}

\usepackage{tikz}
\usepackage{comment}
\usepackage{amsmath,amssymb} 
\usepackage{color}

\usepackage{multirow}
\usepackage{axessibility}  

\begin{document}
\pagestyle{headings}
\mainmatter
\def\ECCVSubNumber{8}  

\title{Feature-level augmentation to improve robustness of deep neural networks to affine transformations} 


\titlerunning{Feature-level augmentation to improve robustness of deep neural networks}
%
\author{Adrian Sandru\inst{1,2} \and
Mariana-Iuliana Georgescu\inst{1,2} \and
Radu Tudor Ionescu\inst{1,2,3,*}}
\authorrunning{Sandru et al.}
%
\institute{Department of Computer Science, University of Bucharest\\
14 Academiei, Bucharest, Romania\\
\and
SecurifAI\\
21D Mircea Voda, Bucharest, Romania\\
\and
Romanian Young Academy, University of Bucharest\\
90 Panduri Street, Bucharest, Romania\\
*\email{raducu.ionescu@gmail.com}}
\maketitle

\begin{abstract}
Recent studies revealed that convolutional neural networks do not generalize well to small image transformations, e.g.~rotations by a few degrees or translations of a few pixels. To improve the robustness to such transformations, we propose to introduce data augmentation at intermediate layers of the neural architecture, in addition to the common data augmentation applied on the input images. By introducing small perturbations to activation maps (features) at various levels, we develop the capacity of the neural network to cope with such transformations. We conduct experiments on three image classification benchmarks (Tiny ImageNet, Caltech-256 and Food-101), considering two different convolutional architectures (ResNet-18 and DenseNet-121). When compared with two state-of-the-art stabilization methods, the empirical results show that our approach consistently attains the best trade-off between accuracy and mean flip rate.
\end{abstract}

\section{Introduction}
\label{sec:intro}

\begin{figure}[!t]
\begin{center}
\centering
\includegraphics[width=0.52\linewidth]{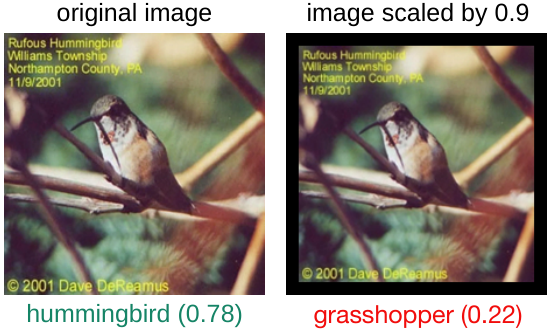}
\vspace{-0.2cm}
\caption{An image of a humming bird from Caltech-256 that is wrongly classified as grasshopper (with a probability of $0.22$) by a ResNet-18 model, after downscaling it by a factor of $0.9$. Best viewed in color.}
\label{fig_problem}
\end{center}
\end{figure}

A series of recent studies \cite{Azulay-JMLR-2019,Chaman-CVPR-2021,goodfellow2014explaining,moosavi2016deepfool,Nicolas-ACM-2017,Szegedy-ICLR-2014,zhang-ICML-2019,Zheng-CVPR-2016} showed that convolutional neural networks (CNNs) are not properly equipped to deal with small image perturbations. Indeed, it appears that a subtle affine transformation, e.g.~a rotation by a few degrees or a translation of a few pixels, can alter the model's decision towards making a wrong prediction. The problem is illustrated by the example shown in Figure~\ref{fig_problem}, where a deep neural model is no longer able to predict the correct class upon downscaling the input image by a factor of $0.9$.
To increase the robustness to such small perturbations, researchers \cite{Azulay-JMLR-2019,Chaman-CVPR-2021,zhang-ICML-2019,Zheng-CVPR-2016} proposed various approaches ranging from architectural changes \cite{Chaman-CVPR-2021,zhang-ICML-2019} and training strategy updates \cite{Zheng-CVPR-2016} to input data augmentations \cite{Michaelis-arxic-2020,Volk-ITSC-2019}. However, to the best of our knowledge, none of the previous works tried to apply augmentations at the intermediate layers of the neural network. We conjecture that introducing feature-level augmentations improves the robustness of deep CNNs to affine transformations. To this end, we present an augmentation technique that randomly selects a convolutional layer at each mini-batch and applies independent affine transformations (translation, rotation, scaling) on each activation map from the selected layer.

To demonstrate the practical utility of our approach, we conduct experiments with ResNet-18 \cite{HE-CVPR-2016} and DenseNet-121 \cite{Gao-CVPR-2017} on three benchmark data sets, namely Tiny ImageNet \cite{Russakovsky-IJCV-2015}, Caltech-256 \cite{Griffin-2007} and Food-101 \cite{Lukas-ECCV-2014}. Importantly, we show that feature-level augmentation helps even when the models are trained with standard data augmentation. When compared with two state-of-the-art methods \cite{Chaman-CVPR-2021,zhang-ICML-2019}, the empirical results show that our approach attains the best trade-off between accuracy and mean flip rate. To the best of our knowledge, this is the first time such a comparison is made.

\paragraph{\bf Contribution.}
In summary, our contribution is twofold:
\begin{itemize}
\item  We introduce a novel method based on feature-level augmentation to increase the robustness of deep neural networks to affine transformations.
\item  We conduct an empirical evaluation study to compare state-of-the-art methods addressing the robustness problem among themselves as well as with our approach.
\end{itemize}

\section{Related work}
\label{sec:relatedart}

In literature, there are several approaches towards improving the robustness of deep neural networks to image perturbations.
A popular and natural strategy, that proved to work sufficiently well, is to train the network using augmented images, as suggested in \cite{Azulay-JMLR-2019,Michaelis-arxic-2020,Volk-ITSC-2019}. The intuition behind this approach is to train the model on a wider domain, which can become more similar to the test data. The experiments conducted by the authors suggest that the robustness and performance improve on all scenarios. Due to its popularity, we apply image augmentation to the baseline models employed in our experiments. An extended solution of using augmented images is mentioned in \cite{Zheng-CVPR-2016}, where the authors employed an additional loss function in order to stabilize the output features of the network in such a manner that a strongly perturbed image should have a similar outcome to the original one. Data augmentation methods are considered to be constrained by the photographers' bias \cite{Azulay-JMLR-2019}. Thus, the model may only learn to generalize to images from a certain (biased) distribution.


One of the recent architectural changes leading to improvements in the stability of CNNs was introduced by Zhang \cite{zhang-ICML-2019}. The observations of the author centered on the fact that modern CNNs are not shift invariant. Thus, small linear changes applied on the input image may have a negative impact on the final result. The source of this issue is considered to be represented by the downsampling process that usually occurs inside neural networks through pooling operations, which breaks the shift-equivariance. A straightforward solution to solve this issue is to avoid subsampling \cite{Azulay-JMLR-2019}, but this comes with a great computational burden. Therefore, Zhang \cite{zhang-ICML-2019} provides adjustments to conventional operations by including a blur kernel in order to reduce the judder caused by downsampling. We conduct experiments showing that this method can deteriorate the quality of the features and negatively impact the final accuracy of the model, despite improving its stability.

Another new architectural design was proposed in \cite{Chaman-CVPR-2021}, where the authors addressed the downsampling issue by proposing a pooling operation that considers all possible grids and selects the component with the highest norm. Their approach was benchmarked against circular shifts of the input, which do not naturally occur in practical scenarios. Hence, we extend their evaluation to generic (non-circular) affine transformations and compare their approach to our own procedure, showcasing the superiority of our method.

Although researchers explored multiple methods to improve the stability of neural models to image perturbations, it seems that there is no technique that can guarantee a never-failing solution. Hence, we consider that addressing the stability problem with better solutions is of great interest to the computer vision community. To the best of our knowledge, we are the first to propose feature-level augmentation as an enhancement to the stability of neural networks.

\section{Method}
\label{sec:method}

\begin{figure}[!t]
\begin{center}
\centering
\includegraphics[width=1.0\linewidth]{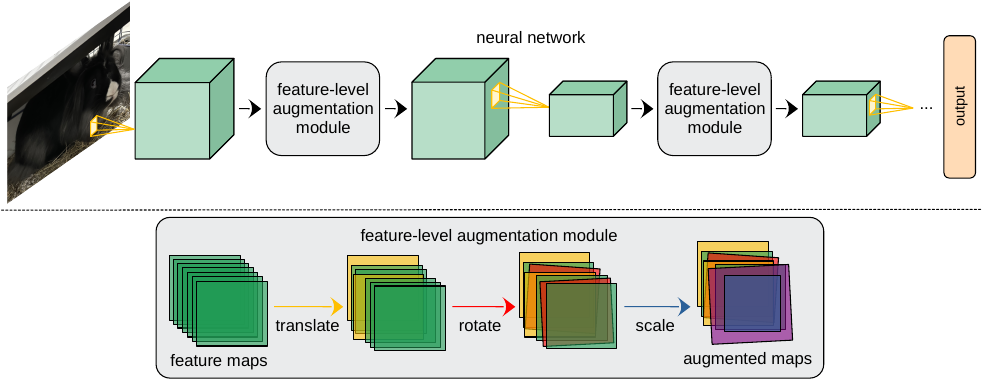}
\vspace{-0.2cm}
\caption{Our feature-level augmentation module is inserted at different levels in the neural architecture. The proposed feature-level augmentation module individually applies the translation, rotation and scaling operations on each activation (feature) map, with a given probability. Best viewed in color.}
\label{fig_pipeline}
\end{center}
\end{figure}

The proposed method consists of extending the conventional input augmentation procedure by applying it at the feature level. In all our experiments, we include the conventional input augmentation, which is based on randomly shifting the images on both axes with values between $-15$ and $+15$ pixels, rotating them by $-15$ to $+15$ degrees, or rescaling them with a factor between $0.4$ and $1.15$, following \cite{Hendrycks-ICLR-2019}. On top of this, we introduce feature-level augmentation (FLA).

We underline that CNNs are usually composed of multiple blocks intercalated with downscaling (pooling) operations. Starting from this observation, we capture the features between two randomly selected consecutive blocks, augment the activation maps provided by the first selected block, and train the network from the second block to the output using an augmented version of the activation maps, as shown in Figure~\ref{fig_pipeline}. 

The augmentation that we propose to employ on the features is based on shifting, rotating and rescaling the activation maps. Due to the fact that one element of a feature map, somewhere deep in the network, is actually the result of processing multiple pixels from the corresponding location in the input, our augmentation procedure should take into account the depth where it is applied, so that the receptive field is not severely affected. Thus, the maximum translation value is scaled accordingly, starting from $15$ pixels near the input, gradually going down to $1$ pixel for the high-level layers. For rotations, we consider a random value between $-15$ and $+15$ degrees, while for the scaling operation, we employ a random resize factor between $0.85$ and $1.15$. In order to avoid noisy features at the beginning, we activate our augmentation procedure only after two training epochs. Afterwards, we apply feature-level augmentation with a probability of $0.5$ on each mini-batch. To increase variety, a feature map has an equal probability of being transformed with any FLA operation. Hence, combining FLA operations is possible.

\section{Experiments}
\label{sec_experiments}

\subsection{Data Sets}

\paragraph{\bf Tiny ImageNet.} The Tiny ImageNet data set is a subset of ImageNet~\cite{Russakovsky-IJCV-2015} containing $120,\!000$ natural images belonging to $200$ classes. The resolution of each image is $64\times64$ pixels. The training set contains $100,\!000$ images, while the validation and test sets contain $10,\!000$ samples each.

\paragraph{\bf Caltech-256.} The Caltech-256~\cite{Griffin-2007} data set contains $30,\!607$ images of $256$ categories. For each object category, we divide the images into $40$ for training, $20$ for validation, leaving the remainder (at least $20$ samples from each category) for the testing stage.

\paragraph{\bf Food-101.} The Food-101~\cite{Lukas-ECCV-2014} data set is formed of $101,\!000$ images of $101$ food types. The original split contains $750$ training images and $250$ test images for each category. We keep $250$ training samples per category to validate the models, leaving us with $500$ samples per category for training.

\subsection{Evaluation Setup}
 
\paragraph{\bf Evaluation measures.} 
For the evaluation, we first employ the accuracy between the ground-truth labels and the predictions of the neural models. Following~\cite{zhang-ICML-2019}, we also use the mean flip rate (mFR) to measure the stability of the models to affine transformations (scaling, rotation and translation). The mFR is measured by how often the predicted label changes, on average, in images with consecutive perturbations. We run each neural model for $5$ times, reporting the average scores and the corresponding standard deviations.
 
Following~\cite{Hendrycks-ICLR-2019}, we create the perturbed version of each test set. We perturb the original test data with rotation, scaling and translation operations. We perform each operation individually on each sample. We rotate the images with angles starting from $-15$ to $+15$ degrees, using a step of $1$ degree. We translate the images by up to $20$ pixels in each direction, using a step of $1$ pixel. We scale the samples by a scaling factor between $0.4$ and $1.15$, using a step of $0.025$.

In order to quantify the trade-off between the accuracy level of a model and its stability to affine transformations, we define the \textit{trade-off} $T$ as:
\begin{equation}\label{eq_total_loss} 
T = \mbox{accuracy} - \mbox{average}_{op}(\mbox{mFR}_{op}),
\end{equation}
where $op \in [\mbox{rotate}, \mbox{scale}, \mbox{translate}]$. A higher value for $T$ represents a better trade-off.

\paragraph{\bf Baselines.}  
As neural models, we choose two very widely used architectures, namely ResNet-18~\cite{HE-CVPR-2016} and DenseNet-121~\cite{Gao-CVPR-2017}. We train the models using common augmentations applied on the input images, such as random shift, scaling and rotation. This represents our first baseline. In addition, we consider two state-of-the-art methods as baselines for improving the stability of these models, namely anti-aliasing (BlurPool)~\cite{zhang-ICML-2019} and adaptive polyphase sampling (APS)~\cite{Chaman-CVPR-2021}. For the BlurPool~\cite{zhang-ICML-2019} method, we used the Triangle-3 filter, which obtains the best trade-off between accuracy and stability to affine transformations. We apply the same input augmentations for all baseline models, as well as for our own models based on feature-level augmentation (FLA). 

\paragraph{\bf Hyperparameter tuning.} We train each model for a maximum of $100$ epochs, halting the training when the value of the validation loss does not decrease for $10$ consecutive epochs. We set the batch size to $16$ samples and the learning rate to $5 \cdot 10^{-4}$. We optimize the models using Adam~\cite{Kingma-ICLR-2015}, keeping the default values for the parameters of Adam.


\begin{table}[!t]
\scriptsize{
\caption{Accuracy scores and mFR values (in $\%$) for translation, rotation and scaling operations on the Tiny ImageNet~\cite{Russakovsky-IJCV-2015}, Caltech-256~\cite{Griffin-2007} and Food-101~\cite{Lukas-ECCV-2014} data sets. Reported results represent the average and the standard deviation over $5$ runs. $\uparrow$ indicates higher values are better. $\downarrow$ indicates lower values are better. Best scores on each data set are highlighted in bold.}
\label{tab_Results}
\begin{center}
\renewcommand{\arraystretch}{1.2}
\begin{tabular}{|c|c|l|c|c|c|c|c|}
\hline
{\bf Data}  & \multirow{2}{*}{\bf Model}               & \multirow{2}{*}{\bf Method} &  \multirow{2}{*}{\bf Accuracy $\uparrow$}    &  \multicolumn{3}{|c|}{\bf mFR $\downarrow$}    & \multirow{2}{*}{\bf Trade-Off $\uparrow$} \\
\cline{5-7}
{\bf Set}  &  &  &  &{\bf Translate}  & {\bf Rotate}  & {\bf Scale} &  \\ 
\hline
\hline
\multirow{8}{*}{\rotatebox{90}{Tiny ImageNet}} &  &  baseline   &   $71.50\pm0.20$    & $12.94\pm0.14$	&  $15.71\pm0.40$ &  $22.29\pm0.21$ & $54.52$\\ 
\cline{3-8}
               &     ResNet  & BP-3~\cite{zhang-ICML-2019}       &   $71.16\pm0.26$    & $\mathbf{10.61\pm0.33}$  &  $14.98\pm0.20$ &  $21.28\pm0.25$ & $55.53$\\
\cline{3-8}
                &   18        & APS~\cite{Chaman-CVPR-2021}         &   $70.30\pm0.22$    & $15.32\pm0.38$  &  $22.91\pm0.23$ &  $27.86\pm0.35$ & $48.27$ \\
\cline{3-8}
                 &           & FLA (ours)         &   $\mathbf{71.76\pm0.16}$    & $12.00\pm0.14$  &  $\mathbf{14.42\pm0.15}$ &  $\mathbf{21.11\pm0.15}$ & $\mathbf{55.91}$\\

\cline{2-8}
                 & &  baseline &   $76.50\pm0.37$  & $9.11\pm0.08$ & $14.19\pm0.33$ & $18.62\pm0.19$ & $62.52$\\ 
\cline{3-8}
                &  DenseNet         & BP-3~\cite{zhang-ICML-2019}       &   $76.57\pm0.24$	& $\mathbf{7.70\pm0.28}$	 &   $13.07\pm0.39$ &  $17.68\pm0.30$ & $63.75$\\
\cline{3-8}
                &  121         & APS~\cite{Chaman-CVPR-2021}         &   $76.00\pm0.21$	& $10.03\pm0.09$	 &   $17.77\pm1.01$ &  $21.29\pm0.26$ & $59.63$\\
\cline{3-8}
                &           & FLA  (ours)       &   $\mathbf{76.60\pm0.44}$	& $8.54\pm0.18$	 &   $\mathbf{12.59\pm0.20}$ &  $\mathbf{17.27\pm0.25}$ & $\mathbf{63.80}$ \\
\hline
\hline

\multirow{8}{*}{\rotatebox{90}{Caltech-256}} &  &  baseline   & $\mathbf{78.96\pm0.30}$  & $5.82\pm0.15$ &	$6.16\pm0.13$ & 	$10.07\pm0.09$ & $71.61$ \\ 
\cline{3-8}
                &  ResNet         & BP-3~\cite{zhang-ICML-2019}       &   $76.92\pm0.26$    & $\mathbf{4.61\pm0.14}$	 &  $\mathbf{4.83\pm0.16}$  & $\mathbf{8.40\pm0.19}$ & $70.97$ \\
\cline{3-8}
                &     18      & APS~\cite{Chaman-CVPR-2021}         &   $78.12\pm0.40$    &  $8.79\pm0.15$ &  $9.86\pm0.11$  &  $13.87\pm0.10$ & $67.28$ \\
\cline{3-8}
                &           & FLA (ours)        &   $78.91\pm0.06$    &  $4.99\pm0.12$ & 	$5.45\pm0.07$  &  $9.24\pm0.05$ & $\mathbf{72.35}$ \\

\cline{2-8} 
                &  &  baseline &  $\mathbf{83.98\pm0.14}$    & $4.00\pm0.13$  &  $4.28\pm0.04$  &  $7.11\pm0.03$ & $78.85$ \\ 
\cline{3-8}
                &    DenseNet       & BP-3~\cite{zhang-ICML-2019}       &  $82.91\pm0.32$	& $\mathbf{3.26\pm0.32}$  & 	$\mathbf{3.52\pm0.07}$  & $\mathbf{6.17\pm0.05}$ & $78.59$ \\
\cline{3-8}
                &    121       & APS~\cite{Chaman-CVPR-2021}        &  $83.53\pm0.11$ & $4.57\pm0.08$	 &  $5.57\pm0.05$  & $8.63\pm0.09$ & $77.27$\\
\cline{3-8}
                &           & FLA (ours)        &  $83.54\pm0.27$	& $3.32\pm0.11$  & 	$3.67\pm0.08$  & $6.31\pm0.13$ & $\mathbf{79.10}$ \\
\hline
\hline 

\multirow{8}{*}{\rotatebox{90}{Food-101}} &  &  baseline   & $\mathbf{76.29\pm0.41}$ & $6.41\pm0.03$ & $7.36\pm0.06$ &	$12.39\pm0.05$ & $67.57$\\
\cline{3-8}
                &  ResNet         & BP-3~\cite{zhang-ICML-2019}       &  $74.38\pm0.24$	 & $\mathbf{4.77\pm0.11}$ &  $\mathbf{5.71\pm0.09}$  & $\mathbf{9.90\pm0.09}$ & $67.58$  \\
\cline{3-8}
                &     18      & APS~\cite{Chaman-CVPR-2021}        &  $76.03\pm0.20$   & $8.77\pm0.21$   & 	$11.41\pm0.21$ 	& $16.29\pm0.08$ & $63.87$\\
\cline{3-8}
                &           & FLA (ours)        &  $76.28\pm0.33$  & $5.69\pm0.02$ &  $6.63\pm0.02$  &  $11.41\pm0.04$ & $\textbf{68.37}$ \\

\cline{2-8}
                &  &  baseline  &   $\mathbf{83.26\pm0.20}$  & $4.08\pm0.07$   &   $4.85\pm0.08$ & $8.31\pm0.08$ & $77.51$\\ 
\cline{3-8}
                &  DenseNet         & BP-3~\cite{zhang-ICML-2019}       &  $81.94\pm0.26$  & $\mathbf{3.26\pm0.05}$ & $\mathbf{4.00\pm0.03}$   & $\mathbf{7.05\pm0.09}$ & $77.17$ \\
\cline{3-8}
                &  121         & APS~\cite{Chaman-CVPR-2021}         &  $82.77\pm0.16$  & $4.69\pm0.04$ & $6.19\pm0.08$   & $10.20\pm0.06$ & $75.74$\\
\cline{3-8}
                &           & FLA (ours)        &   $82.86\pm0.20$ & $3.46\pm0.07$ & $4.21\pm0.10$   &  $7.58\pm0.09$ & $\mathbf{77.78}$ \\
\hline

\end{tabular}
\end{center}
}
\end{table}

\subsection{Results}

We present the results on the Tiny ImageNet~\cite{Russakovsky-IJCV-2015}, Caltech-256~\cite{Griffin-2007} and Food-101~\cite{Lukas-ECCV-2014} data sets in Table~\ref{tab_Results}. As mentioned earlier, we compare our approach with a baseline based on standard data augmentation as well as two state-of-the-art methods, namely APS~\cite{Chaman-CVPR-2021} and BlurPool~\cite{zhang-ICML-2019}. 




\begin{figure}[!t]
\begin{center}
\centering
\includegraphics[width=0.72\linewidth]{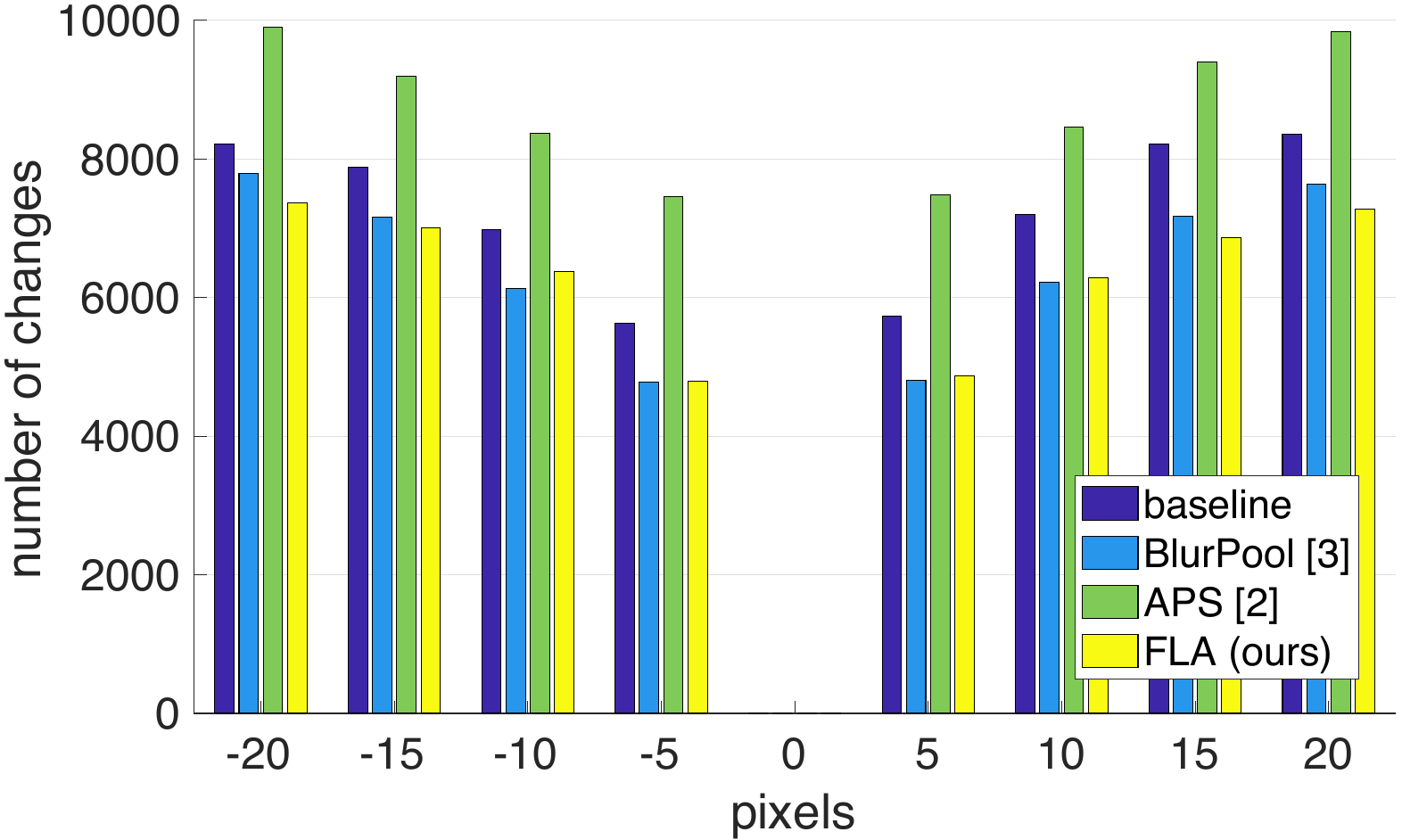}
\vspace{-0.2cm}
\caption{The number of times the labels predicted by a ResNet-18 model having augmented images as input are different from the labels predicted for the original images (translated with $0$ pixels) from Caltech-256. The images are translated from $-20$ pixels to $+20$ pixels. The ResNet-18 model is trained using various methods addressing the stability problem: input data augmentation (baseline), APS~\cite{Chaman-CVPR-2021}, BlurPool~\cite{zhang-ICML-2019} and FLA (ours). Best viewed in color.}
\label{fig_hist}
\end{center}
 
\end{figure}

First, we observe that the APS~\cite{Chaman-CVPR-2021} method does not surpass the accuracy level obtained by the baseline, regardless of the architecture or the data set. It also does not increase the models' stability to affine transformations, even obtaining worse results than the baseline in terms of mFR. The results of APS demonstrate that the $100\%$ stability to circular shift claimed by Chaman et al.~\cite{Chaman-CVPR-2021} does not increase the network's stability to generic affine transformations. 

The BlurPool~\cite{zhang-ICML-2019} method increases the models' stability to affine transformations, performing better than the baseline in terms of mFR. However, it also decreases the accuracy of the model in most cases. The only case when BlurPool increases the accuracy level is on Tiny ImageNet for the DenseNet-121 architecture (the baseline reaches an accuracy of $76.50\%$, while BlurPool reaches a higher accuracy of $76.57\%$). With one exception (the Tiny ImageNet data set), the models trained with the BlurPool method obtain lower trade-off indices compared to the baseline models.

Different from APS~\cite{Chaman-CVPR-2021} and BlurPool~\cite{zhang-ICML-2019}, our method consistently attains the best trade-off between accuracy and stability to affine transformations across all models and data sets, as shown in Table~\ref{tab_Results}. On Tiny ImageNet, we obtain the highest accuracy score and the lowest mFR for rotation and scaling, regardless of the network architecture. On Caltech-256, our method obtains the closest accuracy scores to the baselines, while also increasing the models' stability in terms of all mFR scores. Since our method increases the stability of ResNet-18 during multiple runs on Caltech-256, it seems to have an interesting effect on the standard deviation of the reported accuracy, reducing it from $0.30$ to $0.06$.
On the Food-101 benchmark, our approach applied on the ResNet-18 architecture attains an accuracy level on par with the baseline ($76.28\%\pm0.33$ vs.~$76.29\%\pm0.41$), but the stability of our approach to affine transformations is significantly higher compared to the stability of the baseline ($5.69\%$ vs.~$6.41\%$, $6.63\%$ vs.~$7.36\%$, $11.41\%$ vs.~$12.39\%$ in terms of the mFR for the translation, rotation and scaling operations, respectively).

\begin{figure}[!t]
\begin{center}
\centering
\includegraphics[width=0.8\linewidth]{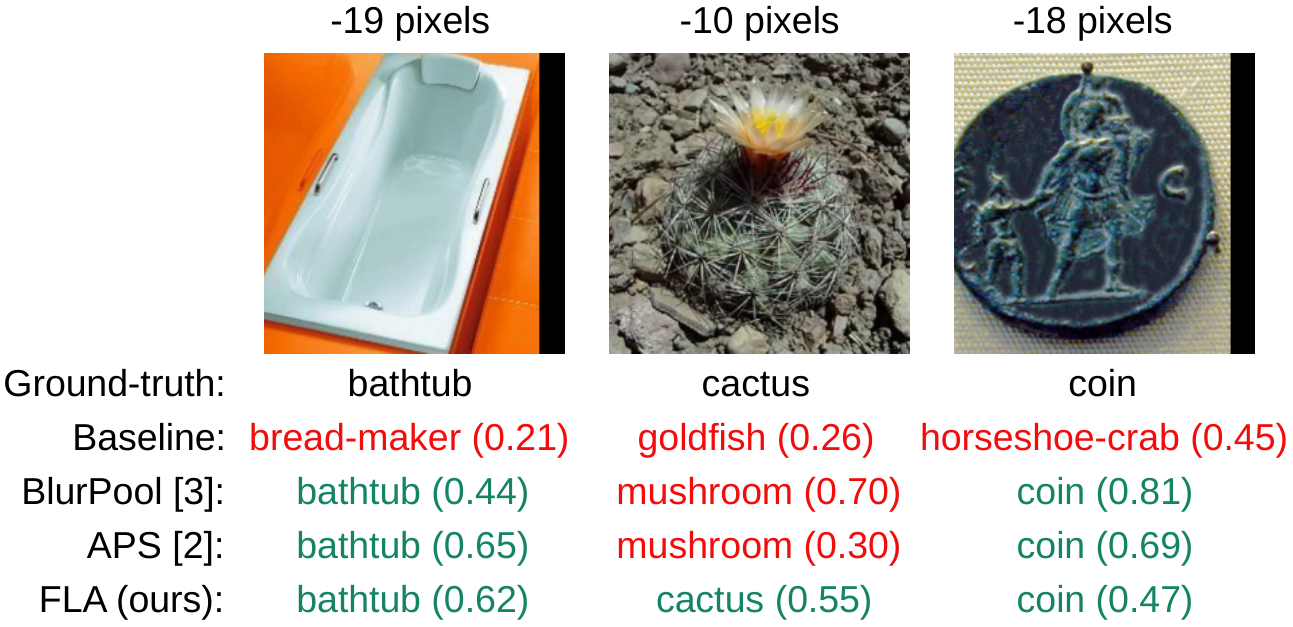}
\caption{Images perturbed with various translations with ground-truth and predicted labels (by the baseline, BlurPool, APS and FLA methods). Best viewed in color.}
\label{fig_samples}
\end{center}
\end{figure}

In Figure~\ref{fig_hist}, we illustrate how many times the labels predicted by a ResNet-18 model having translated images as input are different from the labels predicted for the original images (translated with $0$ pixels). We observe that our method has a significantly smaller number of instabilities (different predictions) than the baseline and APS~\cite{Chaman-CVPR-2021} methods, while the BlurPool~\cite{zhang-ICML-2019} method attains similar results to our method. We also observe that the number of label differences increases with the magnitude of the translation operation for all methods, but the pace seems comparatively slower for our method.

In Figure~\ref{fig_samples}, we show a couple of images from Caltech-256 that are misclassified by the baseline. We observe that BlurPool and APS induce correct labels for two out of three samples, while FLA is able to correct all labels.

\section{Conclusion}
\label{sec:conclusion}

In this paper, we have proposed a novel method to address the stability of CNNs against affine perturbations applied on the input images. To improve stability, our method relies on feature-level augmentation. In addition, we are the first who have conducted comparative experiments to assess the performance of state-of-the-art methods~\cite{Chaman-CVPR-2021,zhang-ICML-2019} addressing the stability problem. 

Although there is a recent trend towards focusing on vision transformers \cite{Dosovitskiy-ICLR-2020,Khan-arXiv-2021,ristea2021cytran,Wu-arXiv-2021}, our study involves only convolutional architectures. However, the recent work of Liu et al.~\cite{liu2022convnet} shows that, upon making proper adjustments, CNNs can obtain comparable results to vision transformers. We thus believe that research related to CNN architectures, such as our own, is still valuable to the computer vision community.

In future work, we aim to study the effect of applying our approach at inference time. This might further improve the stability of neural models, but the stability gains have to be put in balance with the inevitable slowdown in terms of computational time. At the moment, our technique does not affect the inference time at all.

\paragraph{\bf Acknowledgment.}
This article has benefited from the support of the Romanian Young Academy, which is funded by Stiftung Mercator and the Alexander von Humboldt Foundation for the period 2020-2022.

\bibliographystyle{splncs04}
\bibliography{references}

\begin{thebibliography}{10}
\providecommand{\url}[1]{\texttt{#1}}
\providecommand{\urlprefix}{URL }
\providecommand{\doi}[1]{https://doi.org/#1}

\bibitem{Azulay-JMLR-2019}
Azulay, A., Weiss, Y.: Why do deep convolutional networks generalize so poorly
  to small image transformations? Journal of Machine Learning Research
  \textbf{20},  1--25 (2019)

\bibitem{Lukas-ECCV-2014}
Bossard, L., Guillaumin, M., Van~Gool, L.: {Food-101 -- Mining Discriminative
  Components with Random Forests}. In: Proceedings of ECCV. pp. 446--461 (2014)

\bibitem{Chaman-CVPR-2021}
Chaman, A., Dokmanic, I.: {Truly Shift-Invariant Convolutional Neural
  Networks}. In: Proceedings of CVPR. pp. 3773--3783 (2021)

\bibitem{Dosovitskiy-ICLR-2020}
Dosovitskiy, A., Beyer, L., Kolesnikov, A., Weissenborn, D., Zhai, X.,
  Unterthiner, T., Dehghani, M., Minderer, M., Heigold, G., Gelly, S., et~al.:
  An image is worth 16x16 words: Transformers for image recognition at scale.
  In: Proceedings of ICLR (2021)

\bibitem{goodfellow2014explaining}
Goodfellow, I.J., Shlens, J., Szegedy, C.: Explaining and harnessing
  adversarial examples. In: Proceedings of ICLR (2015)

\bibitem{Griffin-2007}
Griffin, G., Holub, A., Perona, P.: {Caltech-256 Object Category Dataset}.
  Tech. rep., California Institute of Technology (2007)

\bibitem{HE-CVPR-2016}
He, K., Zhang, X., Ren, S., Sun, J.: {Deep Residual Learning for Image
  Recognition}. In: Proceedings of CVPR. pp. 770--778 (2016)

\bibitem{Hendrycks-ICLR-2019}
Hendrycks, D., Dietterich, T.: Benchmarking neural network robustness to common
  corruptions and perturbations. Proceedings of ICLR  (2019)

\bibitem{Gao-CVPR-2017}
Huang, G., Liu, Z., Van Der~Maaten, L., Weinberger, K.Q.: {Densely Connected
  Convolutional Networks}. In: Proceedings of CVPR. pp. 2261--2269 (2017)

\bibitem{Khan-arXiv-2021}
Khan, S., Naseer, M., Hayat, M., Zamir, S.W., Khan, F.S., Shah, M.:
  {Transformers in Vision: A Survey}. arXiv preprint arXiv:2101.01169  (2021)

\bibitem{Kingma-ICLR-2015}
Kingma, D.P., Ba, J.: Adam: A method for stochastic optimization. In:
  Proceedings of ICLR (2015)

\bibitem{liu2022convnet}
Liu, Z., Mao, H., Wu, C.Y., Feichtenhofer, C., Darrell, T., Xie, S.: {A ConvNet
  for the 2020s}. arXiv preprint arXiv:2201.03545  (2022)

\bibitem{Michaelis-arxic-2020}
Michaelis, C., Mitzkus, B., Geirhos, R., Rusak, E., Bringmann, O., Ecker, A.S.,
  Bethge, M., Brendel, W.: Benchmarking robustness in object detection:
  Autonomous driving when winter is coming. arXiv preprint arXiv:1907.07484
  (2020)

\bibitem{moosavi2016deepfool}
Moosavi-Dezfooli, S.M., Fawzi, A., Frossard, P.: {DeepFool: a simple and
  accurate method to fool deep neural networks}. In: Proceedings of CVPR. pp.
  2574--2582 (2016)

\bibitem{Nicolas-ACM-2017}
Papernot, N., McDaniel, P., Goodfellow, I., Jha, S., Celik, Z.B., Swami, A.:
  Practical black-box attacks against machine learning. In: Proceedings of ASIA
  CCS. pp. 506--519 (2017)

\bibitem{ristea2021cytran}
Ristea, N.C., Miron, A.I., Savencu, O., Georgescu, M.I., Verga, N., Khan, F.S.,
  Ionescu, R.T.: {CyTran: Cycle-Consistent Transformers for Non-Contrast to
  Contrast CT Translation}. arXiv preprint arXiv:2110.06400  (2021)

\bibitem{Russakovsky-IJCV-2015}
Russakovsky, O., Deng, J., Su, H., Krause, J., Satheesh, S., Ma, S., Huang, Z.,
  Karpathy, A., Khosla, A., Bernstein, M., Berg, A.C., Fei-Fei, L.: {ImageNet
  Large Scale Visual Recognition Challenge}. International Journal of Computer
  Vision  \textbf{115}(3),  211--252 (2015)

\bibitem{Szegedy-ICLR-2014}
Szegedy, C., Zaremba, W., Sutskever, I., Bruna, J., Erhan, D., Goodfellow, I.,
  Fergus, R.: Intriguing properties of neural networks. In: Proceedings of ICLR
  (2014)

\bibitem{Volk-ITSC-2019}
Volk, G., Müller, S., Bernuth, A.v., Hospach, D., Bringmann, O.: {Towards
  Robust CNN-based Object Detection through Augmentation with Synthetic Rain
  Variations}. In: Proceedings of ITSC. pp. 285--292 (2019)

\bibitem{Wu-arXiv-2021}
Wu, H., Xiao, B., Codella, N., Liu, M., Dai, X., Yuan, L., Zhang, L.: {CvT:
  Introducing Convolutions to Vision Transformers}. arXiv preprint
  arXiv:2103.15808  (2021)

\bibitem{zhang-ICML-2019}
Zhang, R.: {Making Convolutional Networks Shift-Invariant Again}. In:
  Proceedings of ICML. vol.~97, pp. 7324--7334 (2019)

\bibitem{Zheng-CVPR-2016}
Zheng, S., Song, Y., Leung, T., Goodfellow, I.: {Improving the Robustness of
  Deep Neural Networks via Stability Training}. In: Proceedings of CVPR. pp.
  4480--4488 (2016)

\end{thebibliography}
\end{document}